\documentclass[sn-mathphys,Numbered]{sn-jnl}

\usepackage{graphicx}%
\usepackage{multirow}%
\usepackage{amsmath,amssymb,amsfonts}%
\usepackage[utf8]{inputenc}

\usepackage{amsthm}%
\usepackage{mathrsfs}%
\usepackage[title]{appendix}%
\usepackage{xcolor}%
\usepackage{textcomp}%
\usepackage{manyfoot}%
\usepackage{multirow}
\usepackage{graphicx}

\usepackage{booktabs}
\usepackage{changepage}

\usepackage{algorithm}%
\usepackage{algorithmicx}%
\usepackage{algpseudocode}%
\usepackage{listings}%
\usepackage{booktabs}
\usepackage{colortbl}
\usepackage{graphicx}
\usepackage{adjustbox}
\usepackage{tabularx}
\usepackage{changepage}

\usepackage{float}
\usepackage{subcaption}
\usepackage{hyperref}
\usepackage{geometry} 
\begin{document}

\title[Article Title]{Boosting House Price Estimations with Multi-Head Gated Attention}

\author*[1]{\fnm{Zakaria Abdellah} \sur{Sellam}}\email{abdellah.sellam@isasi.cnr.it}
\author[1]{\fnm{Cosimo} \sur{Distante}}\email{cosimo.distante@cnr.it}
\author[2]{\fnm{Abdelmalik} \sur{Taleb-Ahmed}}\email{Abdelmalik.Taleb-Ahmed@uphf.fr}
\author[1]{\fnm{Pier Luigi} \sur{Mazzeo}}\email{pierluigi.mazzeo@cnr.it}

\affil*[1]{\orgdiv{Institute of Applied Sciences and Intelligent Systems "Eduardo Caianiello"}, \orgname{CNR}, \orgaddress{\street{}, \city{Lecce},\country{Italy}}}
\affil[2]{\orgdiv{Laboratory of IEMN, CNRS, Centrale Lille, UMR 8520, Univ. Polytechnique Hauts-de-France}, \orgname{Universite Polytechnique Hauts de France}, \orgaddress{\city{F-59313, Valencienne}, \country{France}}}

\abstract{Evaluating house prices is crucial for various stakeholders, including homeowners, investors, and policymakers. However, traditional spatial interpolation methods have limitations in capturing the complex spatial relationships that affect property values. 
To address these challenges, we have developed a new method called Multi-Head Gated Attention for spatial interpolation. Our approach builds upon attention-based interpolation models and incorporates multiple attention heads and gating mechanisms to capture spatial dependencies and contextual information better. Importantly, our model produces embeddings that reduce the dimensionality of the data, enabling simpler models like linear regression to outperform complex ensembling models. We conducted extensive experiments to compare our model with baseline methods and the original attention-based interpolation model. The results show a significant improvement in the accuracy of house price predictions, validating the effectiveness of our approach. This research advances the field of spatial interpolation and provides a robust tool for more precise house price evaluation.
Our GitHub repository.\footnote{\href{https://www.github.com/ldb0071/Final\_file/tree/main/ASI-main}{Final\_file/tree/main/ASI-main}} contains the data and code for all datasets, which are available for researchers and practitioners interested in replicating or building upon our work.}




\maketitle

\section{Introduction}
The Real Estate sector plays a pivotal role in the global economy, with house prices significantly influencing individual wealth and broader economic trends. Fluctuations in house prices can stimulate consumption and boost the economy when prices rise. At the same time, a decrease can limit an individual's borrowing capacity, potentially crowding out investments due to the evaporation in the value of collaterals \cite{case2000}. The shock in the global economy caused by the 2008 housing bubble perfectly illustrates the importance of a stable and measurable house price \cite{reinhart2010}. 
Predicting house prices, however, is a complex task due to the multitude of influencing factors. 
Historically, house price prediction has relied on traditional regression models that consider a range of property-specific factors such as size, age, condition, and number of rooms, among others \cite{bourassa2003}. However, with the advent of machine learning, the landscape of house price prediction has significantly evolved. Techniques such as support vector machines, decision trees, and neural networks have been employed to improve prediction accuracy \cite{chen2019}. In addition, ensemble learning methods, such as boosting, have been used to enhance the performance of prediction models. Specifically, XGBoost \cite{chen2016xgboost}, a scalable and accurate implementation of gradient boosting machines, has been applied to house price prediction with promising results \cite{ajrcos2023v16i2339, bcpbm.v32i.2881}. 
Furthermore, to account for spatial heterogeneity in house prices, Geographically Weighted Regression (GWR) and its variants have been utilised \cite{fotheringham2002, huang2016, wang2018, li2018}. Kriging\footnote{Kriging is a regression method used in spatial analysis (geostatistics) that allows one to interpolate a quantity in space, minimising the mean square error.} \cite{chung2019supplement}, a geostatistical method, has also been used for spatial interpolation in house price prediction \cite{paez2005, kang2017}. 
However, these conventional methodologies bear certain limitations. For instance, they might struggle to capture complex spatial relationships, particularly in regions with diverse and distinct geographical realms. Additionally, assumptions like isotropic variability, which presupposes a constant spatial relationship in all directions, may impede the accuracy of these traditional models in anisotropic landscapes. Moreover, these models might exhibit sensitivity to outliers and could become computationally demanding, especially with increased data points.
Our model builds upon the research of Vianna and Barbosa \cite{viana2021attention}, who developed the attention-based spatial interpolation model. Our research endeavours to extend the paradigm by intertwining Multi-head and Gated Attention mechanisms. Vianna and Barbosa's model manifested a breakthrough by employing an attention mechanism to weigh the influence of neighbouring houses based on supervised learning. They introduced two attention layers: a Euclidean-based attention layer for considering neighbouring houses based on structural feature similarities and a spatial kernel-based attention layer, Geo Attention, for weighing neighbours based on geographic proximity to the target house. These attention layers, coupled with the geographical and structural features of the house, were fed into a fully connected network, culminating in a regression layer for house price prediction. This architecture yielded what they called a 'house embedding,' encapsulating the house attributes and spatial context into a common subspace, serving as a feature set for any regressor to estimate house prices.
Our model extrapolates upon this framework, retaining the essence of generating 'house embeddings' but enhancing the architecture with Multi-head and Gated Attention mechanisms. These innovations are delineated into two distinct attention modules: Geographical Attention and Structural Attention. The Geographical Attention mechanism focuses on spatial relationships and proximities among properties, rendering a more nuanced understanding of the geographical context. Concurrently, the Structural Attention mechanism dives into the intrinsic attributes of properties such as size, age, condition, and the neighbouring points of interest, offering a granular perspective on the structural context.
The Multi-head facet of our model unleashes the potential for parallel processing of geographical and structural information, thereby capturing a rich tapestry of spatial relationships from diverse dimensions. Each head in the Multi-head Attention mechanism could focus on different aspects or scales of spatial relationships, thus enriching the spatial context captured by the model.
Furthermore, the Gated Attention mechanisms are orchestrated to modulate the information flow through the network meticulously. This refined control over the attention distribution is instrumental in mitigating outliers' impact on the estimated values, thereby promising more robust and accurate house price predictions.
Our model, therefore, stands as a sophisticated augmentation of the attention-based spatial interpolation model conceived by Vianna and Barbosa. By synthesising the Multi-head and Gated Attention mechanisms with a bifurcated focus on geographical and structural relationships, our model unfolds a promising avenue for more accurate and insightful real estate price predictions. This innovative approach, rooted in the pioneering work of Vianna and Barbosa, yet elevated with novel attention mechanisms, propels the discourse in this domain towards new vistas, potentially laying a robust foundation for subsequent research endeavours and practical applications in real estate price prediction.
Our project introduces significant contributions to advance real estate price prediction. We combine machine-learning techniques with a deep understanding of spatial heterogeneity in real estate valuations. Below are our contributions:

\begin{itemize}
  \item \textbf{Introducing a New Dataset:} A new dataset for Italian cities has been introduced, likely including features relevant to real estate valuation related to 8 Italian cities.
 
  \item \textbf{Incorporating Various Attention Mechanisms:} applying multi-head gated Attention to capture with different weights and basis to capture different structural and geographical contexts based on the similarities.
  \item \textbf{Testing our model on different datasets:}
  We have tested our approach on other datasets to solidify the model's effectiveness in predicting house prices in other areas and diverse datasets.
  
\end{itemize}
The subsequent sections of this document are structured as follows: Section 2 presents a comprehensive overview of relevant works, including literature and methodologies, that relate to house price estimation and section 3 delves into our proposed attention network, detailing its unique features and potential benefits. In Section 4, we conduct experiments, perform data analysis, and provide a thorough evaluation of our model. Lastly, in Section 5, we draw insightful conclusions based on our experimentation, compare our approach with prior methodologies, and articulate the implications of our findings. This structure ensures a coherent and comprehensive understanding of our innovative methodology for house price prediction.

\section{Related works}\label{sec2}
House price estimation is a critical activity with far-reaching implications in the real estate industry. This field has been the subject of extensive academic research, traditionally employing regression analyses that integrate multiple variables, data types, and methodologies. In this review, we explore the scholarly landscape of this subject, tracing the evolution of research methodologies and spotlighting modern advancements and emerging trends.

The Hedonic Price Theory, first introduced by Rosen in 1974 \cite{Rosen1974}, is the foundation for Hedonic Regression models. These models have become a crucial tool in studying house prices. The theory utilises a set of attributes, such as the number of bedrooms or bathrooms, to explain and represent a house's market value. These attributes are ranked based on their impact on a house's utility function, assuming that a market equilibrium between buyers and sellers determines the sale price. Hedonic Regression models are widely used to analyse the effects of different factors on house prices in various areas, making them a robust tool for market segmentation \cite{Yao2018}. Although the original Hedonic Price Theory focused mainly on the intrinsic characteristics of a house, it has evolved to account for external factors like location \cite{Frew2002}. This adaptation was motivated by the realisation that solely considering a house's intrinsic attributes was insufficient for accurate price representation \cite{Limsombunchai2004}. Despite its widespread use, Hedonic Regression models have faced challenges, including issues related to the stability of attribute coefficients across different locations and property types, as well as limitations in handling non-linearity and model specification \cite{Wang2014}.

The integration of machine learning into house price prediction has been significantly accelerated by advancements in computational capabilities and the increase of data \cite{Chaphalkar2013}. Initially, the focus was mainly on traditional machine learning algorithms such as Linear Regression (LR) \cite{Cook1977}. While these linear models offered computational efficiency and ease of interpretation, they were limited in capturing the high-dimensional and non-linear complexities inherent in transaction price data. To address these limitations, researchers explored regularisation techniques like Ridge and Lasso Regression \cite{tibshirani1996regression, hoerl1970ridge}. These methods helped mitigate overfitting and offered a more refined approach to feature selection but struggled with capturing complex, non-linear relationships. Principal Component Analysis (PCA) \cite{jolliffe1986principal} has also been employed for dimensionality reduction to simplify the feature space, although it has been criticised for potentially discarding crucial information. This led to the exploration of more flexible, non-linear models such as Support Vector Regression (SVR) \cite{Drucker1997SupportVector} and Decision Trees \cite{quinlan1986induction}. Support Vector Regression (SVR) offers a solution for non-linearities through various kernel functions, while Decision Trees provide a simple yet effective approach for detecting non-linear patterns \cite{Drucker1997SupportVector, quinlan1986induction}. However, Decision Trees are prone to overfitting. To combat this, ensemble methods like Random Forests were developed to improve model generalisation \cite{ho1995random}. Random Forests combine the outcomes of many decorrelated trees to minimise variance and enhance accuracy.

With advancements in computational power, the field has shifted to more sophisticated ensemble methods such as XGBoost \cite{Chen2016}. Unlike Random Forests, XGBoost constructs trees sequentially to correct the errors made by the previous ones. This makes XGBoost particularly effective in handling diverse data structures and enhancing prediction accuracy \cite{pavlyshenko2018using}. These advanced ensemble models are also highly scalable and efficient, often surpassing Random Forests' performance on large datasets.

To further optimise their predictive performance, these sophisticated ensemble models are often fine-tuned using metaheuristic optimisation techniques like Particle Swarm Optimization (PSO) \cite{Claesen2015, Alfyatin2017}. These optimisation techniques enable precise tuning of hyperparameters, resulting in models that are both accurate and computationally efficient.

The latest development in house price prediction is Graph Neural Networks (GNNs) \cite{zhou2020graph}, which excel in identifying spatial relationships between properties. However, GNNs can be computationally demanding and require large, well-curated datasets for practical training. Additionally, their performance can vary significantly based on the architecture and hyperparameters, which may hinder their widespread adoption.

Furthermore, the domain has seen the rise of deep learning techniques. Deep Neural Networks (DNNs) \cite{schmidhuber2015deep} can automatically learn feature representations, eliminating the need for manual feature engineering. Although DNNs can unravel highly complex relationships in the data, they present challenges, such as the risk of overfitting and the need for substantial datasets and computational resources for practical training.

Building on these advancements, recent research has focused on integrating diverse computational models and data sources. A groundbreaking study by Tchuente et al. \cite{tchuente2022real} on the French real estate market is a prime example. Utilising machine learning techniques such as Random Forest, AdaBoost \cite{freund1997decision}, and gradient boosting \cite{friedman2001greedy}, along with geocoding features, they analysed five years of historical real estate transactions provided by the French government. Their findings revealed that incorporating geocoding elements increased the models' predictive accuracy by over 50

Building upon the findings of Tchuente et al., the research conducted by Zhao et al. \cite{zhao2022pate} represents a significant advancement in data analysis. By incorporating a multi-modal approach that encompassed traffic patterns, amenities, and social emotions in the bustling city of Beijing, China, this study validated the crucial role of location-based data. Furthermore, it introduced a feature-ranking mechanism that established a direct correlation between the data and its economic impact. This groundbreaking research underscores the potential of geolocated data in predicting real estate prices and highlights its transformative capabilities.
Further advancing this research domain, De Nadai et al. \cite{de2018economic} delved into the economic repercussions of neighbourhood characteristics within Italian urban landscapes. Their investigative toolkit encompassed a rich array of data sources including OpenStreetMap\footnote{\href{https://download.geofabrik.de/europe/italy}{europe/italy.html}}, Urban Atlas 2012, imagery from Google Street View, Italian census data\footnote{\href{https://www.istat.it/}{https://www.istat.it/}}, alongside property tax records sourced from the "Immobiliare. it"\footnote{\href{https://www.immobiliare.it/}{www.immobiliare.it}} platform. Through the application of their model, they witnessed a notable 60\% enhancement in nowcasting housing prices, thereby underpinning the transformative potential of leveraging rich, geolocated datasets.
Sarkar Snigdha Sarathi Das et al. \cite{das2021boosting}
It has introduced the concept of Geospatial Network Embedding (GSNE). Unlike traditional models that often overlook the geospatial context of neighbourhood amenities, GSNE aims to capture this crucial aspect. The study emphasises that the proximity of a house to key points of interest (POIs) like train stations, highly-ranked schools, or shopping centres can significantly influence its price. The GSNE model leverages graph neural networks to create embeddings of houses and various types of POIs in multipartite networks. In these networks, houses and POIs are attributed nodes, representing their relationships as edges. This is particularly promising because it allows the model to understand complex latent interactions between houses and POIs, offering a robust and effective way to incorporate geospatial context.\\
Yuhao Kang et al. 
Kang et al.\cite{kang2021understanding} delve into house price appreciation rates, employing a multi-source extensive geo-data framework that amalgamates structural attributes, locational amenities, and visitor patterns, employing machine learning models and geographically weighted regression for accurate predictions at both micro and macro scales. Their gradient-boosting machines achieve an R-squared value of 74\% at the neighbourhood scale, highlighting the effectiveness of their approach in understanding house price appreciation nuances.
On a similar innovative trajectory, Pei-Ying Wang et al. \cite{9395585}. Propel house price prediction forward by harnessing a Joint Self-Attention Mechanism intertwined with a rich analysis of heterogeneous data, including public facilities and environmental aesthetics captured through satellite imagery. Tested in Taipei and New Taipei, this model eclipses other machine learning-based models in prediction accuracy, showcasing a lower error rate. The Spatial Transformer Network (STN)\cite{Jaderberg2015} and their model's novel joint self-attention mechanism intricately dissect the complex relations between different attributes impacting house prices. This work accentuates the necessity of a holistic data-rich approach and extends the versatility of the attention mechanism across various domains, setting a robust foundation for future research.
In a parallel vein, Viana and Barbosa \cite{viana2021attention} introduce a groundbreaking framework that melds the spatial essence of real estate with the structural attributes of houses. Their hybrid attention mechanism orchestrates a balanced blend between the Euclidean space of structural features and the geographic tapestry, crafting them into a unified predictive model. The inception of a house embedding vector carries through the regression analysis domain, offering a fresh lens to capture spatial dependencies. This attention-infused approach heralds a promising avenue where the convergence of spatial interpolation and machine learning unravels a richer understanding of housing market dynamics, further amplifying the potential of attention mechanisms in elucidating the multifaceted nature of house price predictions.
The related work showcases a trajectory towards crafting more nuanced, robust, and insightful real estate price prediction models. These models progressively harness multi-source, geolocated data and sophisticated machine learning techniques, notably attention mechanisms. This evolution reflects a maturing field poised to address the intricate challenges inherent to urban landscapes and real estate markets.

\section{Methodology}\label{sec3}
Our proposed methodology aims to create robust house embeddings by assessing the similarity between a specific house and its neighbouring properties. This approach goes beyond merely considering individual property attributes and geographical location. Instead, it encapsulates each house's local characteristics with its immediate surroundings. Unlike traditional methods, we integrate the geographical coordinates of the property to refine this embedding further, capturing the essence of its surroundings and their relation to critical landmarks or amenities.

\begin{figure*}
    \centering
    \includegraphics[width=1.1\linewidth]{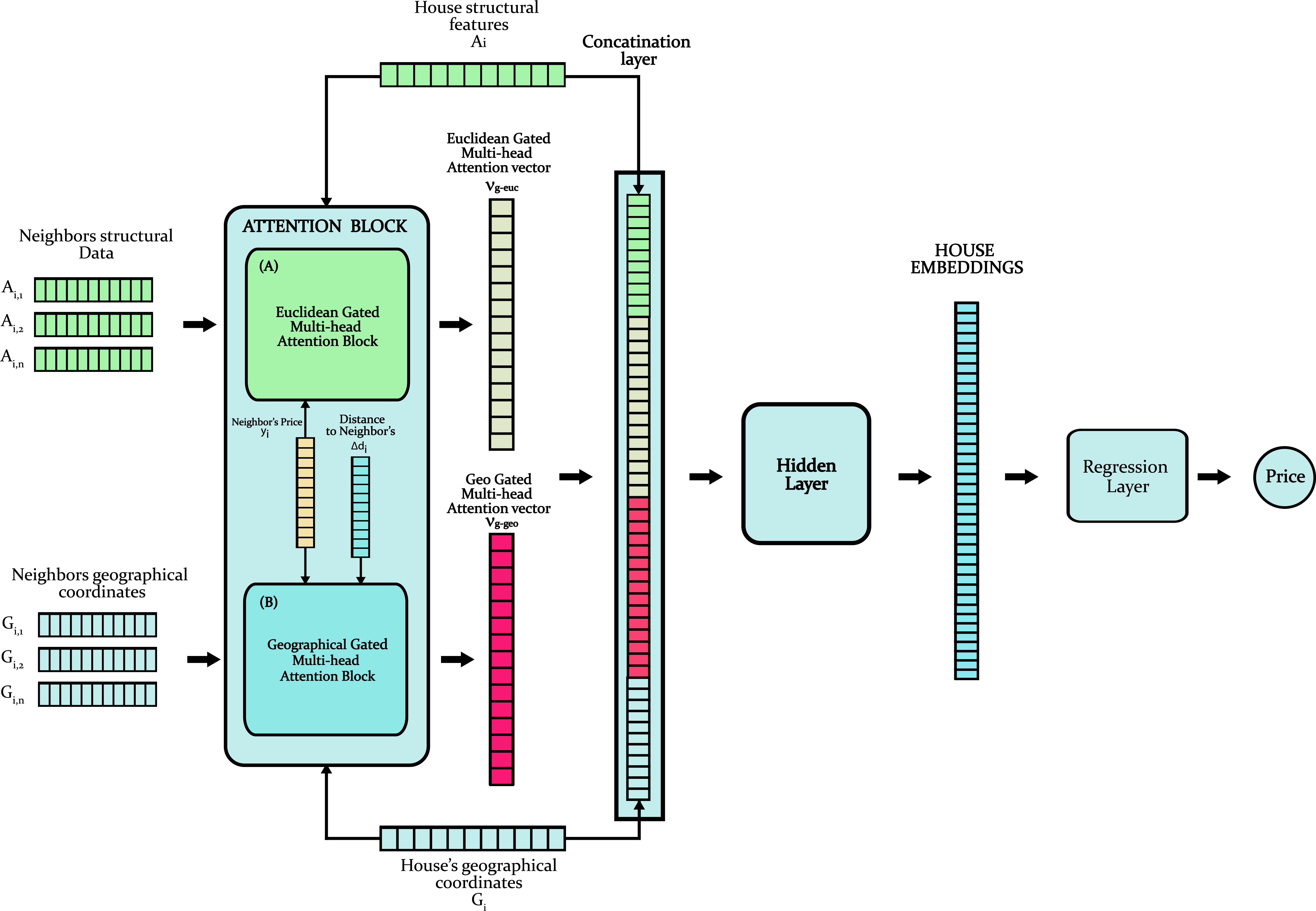}
    \caption{Architecture representation of the multi-head gated-attention-based interpolation. (A) Represent the Euclidean interpolation block based on the multi-head gated Attention.
    (B) Represent the geo-interpolation block based on the Multi-Head Gated attention.}
    \label{fig:mylabel}
\end{figure*}

Our approach is based on the Attention-Based Spatial Interpolation (ASI) architecture proposed by Viana and Barbosa~\cite{viana2021attention}. This architecture creates geographical and Euclidean similarities and emphasises specific similar points using an attention mechanism. However, more than a simple attention head may be required to capture differentiated interrelations. For this reason, our model employs multi-head-gated attention mechanisms to optimise the extraction of these features and their interrelationships.
Multi-Head Gated Attention allows the model to capture multiple contexts, such as architectural styles, proximity to amenities, and other relevant features. Concurrently, the gated attention mechanism controls the flow of information to ensure that only the most pertinent attributes are considered. This is particularly useful when there is a significant variance between the target house and its neighbours, allowing the model to focus on the most critical similarities or differences.
The Euclidean Multi-Head Gated Attention layer, represented in Figure~\ref{fig:mylabel} (A), calculates attention weights for the structural features of neighbouring houses based on their Euclidean distance to \(A_i\). Concurrently, the Geographical Multi-Head Gated Attention layer in Figure~\ref{fig:mylabel} (B) learns the spatial correlations between the n-nearest geographical neighbours of house i. The output vectors from both attention layers are concatenated with \(A_i\) and \(G_i\) and fed into a fully connected neural network, culminating in a regression layer. This architecture synthesises the influence of the neighbouring houses and the target house's attributes into a single vector, termed the ``house embedding'' illustrated in Figure~\ref{fig:mylabel}.

\subsection{Background knowledge}
To perform predictive analysis in real estate valuation, it is crucial to have a solid foundation of knowledge. This field employs a variety of methodologies and algorithms that are based on fundamental principles and metrics. Understanding these concepts is essential for accurately performing advanced analytical techniques. This subsection aims to clarify some of these key concepts and metrics, providing a starting point for a deeper exploration and comprehension of the subsequent methodologies and evaluations.
\subsubsection{Similarity calculation}\label{subsec2}
In the intricate landscape of data science, similarity is a critical underpinning for various algorithms and methodologies. This sub-subsection aims to illuminate the key metrics ubiquitously employed to quantify similarity, laying the groundwork for the following analyses.

\begin{itemize}
    \item \textbf{Euclidean Distance}: A foundational metric in geometry, Euclidean distance provides a straightforward measure of similarity by calculating the straight-line distance between two points in an Euclidean space.
    \begin{equation}
    d(P_1, P_2) = \sqrt{(x_2 - x_1)^2 + (y_2 - y_1)^2}
    \label{eq:euclidean_distance}
    \end{equation}

    \item \textbf{Cosine Similarity}: This metric is invaluable in high-dimensional spaces, measuring the cosine of the angle between two vectors. It is especially pertinent in text analysis and natural language processing.
    \begin{equation}
    \text{Cosine Similarity} = \frac{C \cdot D}{\|C\| \times \|D\|}
    \label{eq:cosine_similarity}
    \end{equation}

    \item \textbf{Jaccard Index}: A set-based metric, the Jaccard Index is helpful for categorical data, quantifying the ratio of the intersection to the union of two sets.
    \begin{equation}
    J(C, D) = \frac{|C \cap D|}{|C \cup D|}
    \label{eq:jaccard}
    \end{equation}

    \item \textbf{Identity Similarity}: This is a binary similarity measure used to ascertain whether or not two data points are identical. Unlike continuous similarity measures, the Identity Similarity scores one if the data points are similar and 0 if they differ. This measure is handy in scenarios requiring exact matching or where data is categorical. Mathematically, it is expressed as:

    \item \textbf{Gaussian Kernel}: Also known as the Radial Basis Function (RBF) with Gaussian form, this metric is a cornerstone in non-linear data transformations. Unlike other metrics that measure distance directly, the Gaussian Kernel calculates similarity by mapping the original data points into a higher-dimensional space through a Gaussian function. This allows it to capture complex, non-linear relationships between data points. Mathematically, it is expressed as:
    \begin{equation}
    K(x, y) = \exp\left(-\frac{\|x - y\|^2}{2\sigma^2}\right)
    \label{eq:gaussian}
    \end{equation}

    The parameter \( \sigma \) controls the spread of the Gaussian function, thereby influencing the similarity measure. A smaller \( \sigma \) will result in a narrower Gaussian function, making the similarity measure more sensitive to the distance between data points.
\end{itemize}

These metrics serve as the backbone for various algorithms and offer a nuanced understanding of how data points relate to each other in complex spaces, with the Gaussian Kernel standing out for its ability to capture non-linear relationships.
\subsubsection{Spatial interpolation}\label{subsec2}
Spatial interpolation is a critical technique for predicting unknown values at unobserved locations based on known values at observed locations, finding applications in diverse fields such as geostatistics, environmental science, and real estate. The effectiveness of spatial interpolation is intrinsically tied to the choice of similarity measures. For instance, Euclidean distance can be employed in a straightforward approach like "inverse distance weighting" (IDW) \cite{shepard1968two}, where the influence of a neighbouring point on the interpolated value is inversely proportional to its Euclidean distance from the target location. On the other hand, the Gaussian Kernel \cite{you2017image} offers a more nuanced approach by transforming the Euclidean distance into a measure of similarity, thereby capturing complex, non-linear spatial relationships. This is especially useful in advanced geostatistical methods like kriging \cite{matheron1969krigeage}. Therefore, the choice between straightforward measures like Euclidean distance and more complex ones like the Gaussian Kernel can significantly impact the quality of spatial interpolation, exemplifying the broader applicability and importance of similarity measures in data science.
\subsubsection{Attention Mechanisms}\label{subsec2}
Attention mechanisms \cite{vaswani2017attention} has emerged as a cornerstone in many deep learning models, predominantly in sequence-to-sequence tasks such as machine translation and speech recognition. The essence of Attention is to emulate the human ability to focus on specific segments of input data, much like how we selectively concentrate on some aspects of a visual scene or a conversation. Among the diverse attention mechanisms, Soft Attention is a mechanism that computes a weighted sum of all input values. These weights, indicative of the relevance of each input, are typically determined using a softmax function, ensuring a normalised distribution where the weights sum up to one. The continuous nature of these weights makes soft Attention inherently differentiable, rendering it particularly amenable to gradient-based optimisation techniques \cite{bahdanau2014neural}.
On the other hand, intricate Attention operates more selectively. Instead of distributing focus across all inputs, it zeroes in on a specific subset, effectively sidelining the others. Given its discrete selection process, traditional backpropagation struggles with optimising intricate Attention. Yet, this challenge is surmountable with techniques like the reinforce algorithm \cite{mnih2014recurrent}. The \textit{Gated Attention} mechanism \cite{zhang2018gaan} bridges the gap between these two. It adeptly amalgamates information from diverse sources and employs gating tools to ascertain the relevance of each source. This approach can be perceived as a harmonious blend of the soft and hard attention paradigms, encapsulating the strengths while mitigating their limitations \cite{luong2015effective}.
\subsection{Attention Block}
\begin{figure}[h]
    \centering
    \includegraphics[width=0.7\linewidth]{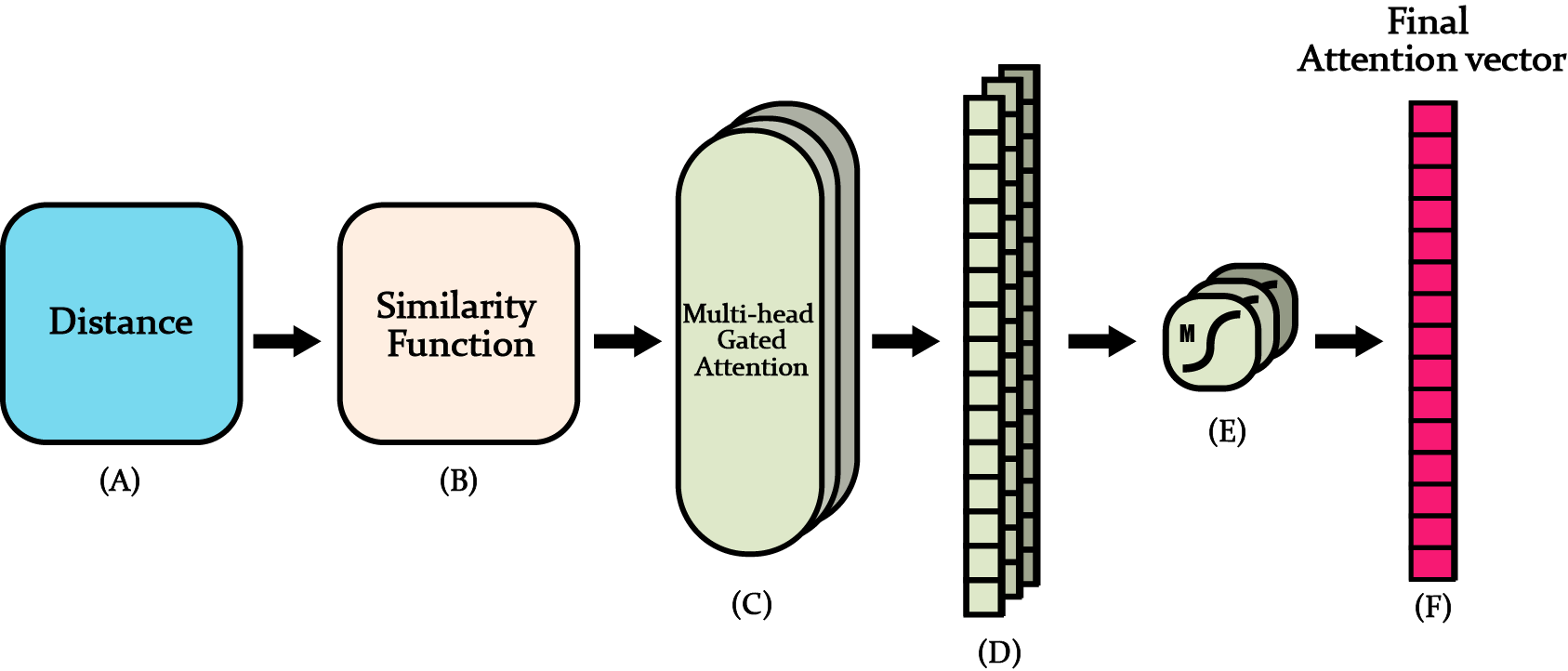}
\caption{Comprehensive overview of the Gated Multi-head Attention mechanism within the Attention Block. (A) depicts the initial computation of geodesic and Euclidean distances, serving as the foundation for subsequent attention calculations. (B) illustrates the Similarity Function, which transforms these foundational distances into similarity scores. (C) shows the core Multi-Head Gated Attention Block, where these similarity scores derive gated attention weights across multiple heads. (D) Highlights the Aggregated Attention Head, consolidating the gated attention weights from all heads into a singular vector. (E) represents the aggregation of multiple gated attentions for each weighted sum. (F) Indicates the Final Attention Vector.}

    \label{fig:multihead}
\end{figure}

The Attention Block is the computational nucleus of our architecture, designed to intricately capture the spatial relationships essential for precise house price prediction. As delineated in Figure~\ref{fig:mylabel}, this block comprises two main components: the Geo Multi-head Gated Attention and the Euclidean Multi-head Gated Attention. Each of these components consists of several key stages, contributing to generating their respective geo- and Euclidean-gated attention vectors. Figure~\ref{fig:multihead} elucidates the fundamental principles for calculating the Geo and Euclidean attention mechanisms. In the initial stage, represented by Figure~\ref{fig:multihead} (A), the Distance Calculation Block computes the distance between the target house and its neighbours. The nature of this distance is contingent on the specific attention mechanism in play, be it Geo or Euclidean. The Similarity Calculation Block, as depicted in Figure~\ref{fig:multihead} (B), transforms these distances into similarity scores. A Gaussian kernel function is employed for Geo Attention, while alternative kernel functions may be used for the Euclidean variant. The subsequent component is the multi-head gated Attention, illustrated in Figure~\ref{fig:multihead} (C). This block leverages the similarity scores to derive attention weights, which are then gated to modulate their influence. The entire process is executed across multiple heads, capturing various facets of the spatial relationships between the target house and its neighbours. Next, the aggregated attention head, represented by Figure~\ref{fig:multihead} (D), consolidates the outputs from all attention heads into a single vector. This is achieved through a weighted sum, where the weights are adaptively learned during training. If the architecture employs multiple attention mechanisms, such as Geo and Euclidean, their aggregated attention heads are combined further. Following this, Figure~\ref{fig:multihead} (E) illustrates the Final Aggregation Block. In this stage, aggregated normalised gating weights are computed using a softmax function. After that, the weighted sums produced from each attention head are multiplied by these normalised weights. This aggregation is performed separately for the Geo and Euclidean attention mechanisms, resulting in their aggregated attention vectors. Finally, the vector produced from this aggregation process is the final attention vector, as depicted in Figure~\ref{fig:multihead} (F). In summary, the Attention Block encapsulates the Multi-head Geo Gated Attention and the Euclidean Multi-head Gated Attention, generating their respective Geo and Euclidean Gated Attention Vectors.

\subsubsection{Geo Multi-head Gated Attention}
The Geo Multi-head, Gated Attention mechanism is intricately designed to capture the spatial relationships between a target house and its neighbouring properties. 
This involves using a Gaussian kernel function to calculate geographic similarity scores between the target house and its neighbours. Equation~\ref{eq:geo_score} demonstrates how the geographic score between the target house \( G_i \) and its neighbouring house \( G_{i,j} \) is computed using the Gaussian kernel function.
\begin{equation}
s(G_i, G_{i,j}) = \exp\left(-\text{geo\_dist}(G_i, G_{i,j}) \times \rho\right)
\label{eq:geo_score}
\end{equation}

Here, \( \rho = \frac{\sigma^2}{2} \) and \( \text{geo\_dist}(G_i, G_{i,j}) \) represents the geodesic distance between \( G_i \) and \( G_{i,j} \).
The vector of similarity scores \( L \) is then transformed into a hidden representation \( H' \) through a fully-connected layer, as described in Equation~\ref{eq:hidden_rep_1}:
\begin{equation}
H' = W' \cdot L + b'
\label{eq:hidden_rep_1}
\end{equation}

In this equation, \( W' \) and \( b' \) are the learned weights and bias terms, respectively.
The attention weights \( a_{\text{geo}} \) are computed using a softmax layer, as formulated in Equation~\ref{eq:attention_weights_1}:
\begin{equation}
a_{\text{geo}}(G_i, G_{i,j}) = \frac{\exp(H'_j)}{\sum_{j'=1}^{n} \exp(H'_{j'})}
\label{eq:attention_weights_1}
\end{equation}

Then, using our defined gated attention mechanism (Equation~\ref{eq:gated_attention}), we apply it to the attention weights:

\begin{equation}
\text{Gate}(x) = \sigma(W_g \cdot x + b_g)
\label{eq:gated_attention}
\end{equation}

Subsequently:

\begin{equation}
a'_{\text{geo}}(G_i, G_{i,j}) = \text{Gate}(a_{\text{geo}}(G_i, G_{i,j})) \odot a_{\text{geo}}(G_i, G_{i,j})
\end{equation}
Where:
\begin{itemize}
    \item \( x \) is the input value, in this case, the original attention weight \( a_{\text{geo}}(G_i, G_{i,j}) \).
    \item \( W_g \) represents the learned weight matrix associated with the gate.
    \item \( b_g \) denotes the bias term.
    \item \( \sigma \) is the sigmoid function, ensuring the output value of the gate lies in the [0,1] range.
\end{itemize}

With this, the Geo Gated Attention Vector \( v_{\text{ggeo}}(G_i) \) is computed as a weighted sum of the features of the neighbouring houses using the modified attention weights \( a'_{\text{geo}} \):

\begin{equation}
v_{\text{ggeo}}(G_i) = \sum_{j=1}^{n} a'_{\text{geo}}(G_i, G_{i,j}) [G_{i,j} \oplus A_{i,j} \oplus \Delta d_{i,j} \oplus y_{i,j}]
\end{equation}

In this equation, \( \Delta d_{i,j} \) represents the geographic distance between house \( i \) and its neighbour \( j \). Similarly, \( y_{i,j} \) signifies the price of the neighbor \( j \), and \( \oplus \) denotes the concatenation operation. The dimensionality of \( v_{\text{geo}}(G_i) \) is derived from the summation of dimensions where \( G_{i,j} \in \mathbb{R}^2 \), \( A_{i,j} \in \mathbb{R}^T \), \( \Delta d_{i,j} \in \mathbb{R}^1 \), and \( y_{i,j} \in \mathbb{R}^1 \). Consequently, the vector \( v_{\text{geo}}(G_i) \) can be viewed as a weighted sum of vectors \( G_{i,j} \), concatenated with \( \Delta d_{i,j} \) and \( y_{i,j} \), and weighted using the normalised geo gated attention coefficients which are determined during the training process.

\subsubsection{Euclidean Multi-head Gated Attention}
The Euclidean Multi-head Gated Attention mechanism is precisely engineered to emphasise the most relevant structural similarities between a target house and its neighbouring properties. This mechanism employs the Euclidean distance to compute the similarity scores between the target house and its neighbours.
The Euclidean distance between the target house \( A_i \) and a neighboring house \( A_{i,j} \) is computed as shown in Equation~\ref{eq:euc_distance}:

\begin{equation}
d(A_i, A_{i,j}) = \sqrt{\sum_{p=1}^{T} (a_{i,p} - a_{i,j,p})^2}
\label{eq:euc_distance}
\end{equation}
where 
\( d(A_i, A_{i,j}) \) is the Euclidean distance indicating similarity between houses based on structural attributes,
\( A_i \) represents the structural features of the target house \( i \),
\( A_{i,j} \) denotes the structural features of the \( j^{th} \) neighboring house to \( i \),
\( a_{i,p} \) and \( a_{i,j,p} \) are specific structural attributes of houses \( i \) and \( j \), respectively, 
and \( T \) is the total number of structural attributes considered.

After computing the Euclidean distances, we construct a vector of similarity scores \( L \), which is then transformed into a hidden representation \( H \) through a fully-connected layer, as described in Equation~\ref{eq:hidden_rep}:

\begin{equation}
H = W \cdot L + b
\label{eq:hidden_rep}
\end{equation}

In Equation~\ref{eq:hidden_rep}, \( W \) and \( b \) are the learned weights and bias terms, respectively.
The attention weights \( a_{\text{euc}} \) are computed using a softmax layer, as formulated in Equation~\ref{eq:attention_weights}:

\begin{equation}
a_{\text{euc}}(A_i, A_{i,j}) = \frac{\exp(H_j)}{\sum_{j'=1}^{n} \exp(H_{j'})}
\label{eq:attention_weights}
\end{equation}

The essence of the gated attention mechanism is to refine the attention weights by introducing an additional modulation step. This modulating factor, or "gate", is typically represented as a value between 0 and 1 and is applied element-wise to the attention weights. The purpose is to amplify or diminish the original attention values based on the model's learned parameters. 

Given this, the gated attention can be defined as:

\begin{equation}
\text{Gate}(x) = \sigma(W_g \cdot x + b_g)
\end{equation}

Where:
\begin{itemize}
    \item \( x \) is the input value, in this case, the original attention weight \( a_{\text{euc}}(A_i, A_{i,j}) \).
    \item \( W_g \) represents the learned weight matrix associated with the gate.
    \item \( b_g \) denotes the bias term.
    \item \( \sigma \) is the sigmoid function, ensuring the output value of the gate lies in the [0,1] range.
\end{itemize}

Subsequently, the gated attention mechanism can be formalised as:

\begin{equation}
a'_{\text{euc}}(A_i, A_{i,j}) = \text{Gate}(a_{\text{euc}}(A_i, A_{i,j})) \odot a_{\text{euc}}(A_i, A_{i,j})
\end{equation}

Here, \( \odot \) denotes element-wise multiplication. Thus, the attention weight is modulated by its gating value, allowing the model to allocate attention more selectively to houses exhibiting the most congruent features.

The Vector with Euclidean Gated Attention denoted as \( v_{\text{geuc}}(A_i) \), represents a cumulative weighted mix of attributes from the surrounding homes. This process uses the gated attention coefficients \( a'_{\text{euc}} \) and is illustrated in Equation~\ref{eq:euc_vector}:

\begin{equation}
v_{\text{geuc}}(A_i) = \sum_{j=1}^{n} a'_{\text{euc}}(A_i, A{i,j}) \odot  [A_{i,j} \oplus y_{i,j}]
\label{eq:euc_vector}
\end{equation}

Within Equation~\ref{eq:euc_vector}, \( y_{i,j} \) defines the price of the \( j^{th} \) neighboring home of house \( i \), while \( \oplus \) denotes the concatenation action. The size of \( v_{\text{euc}}(A_i) \) stands at \( T + 1 \) given that \( A_{i,j} \) resides in \( \mathbb{R}^T \) and \( y_{i,j} \) is part of \( \mathbb{R}^1 \). The composition of \( v_{\text{euc}}(A_i) \) involves initially multiplying the combined vector \( [A_{i,j} \oplus y_{i,j}] \) for each \( j^{th} \) neighbor of house \( i \) by its respective gated attention coefficient \( a'_{\text{euc}}(A_i, A_{i,j}) \), producing an individual weighted vector for every \( j^{th} \) neighbor. An overall summation is then applied to these vectors for all \( n \) adjacent homes to house \( i \). Consequently, the elements within \( v_{\text{euc}}(A_i) \) represent a comprehensive weighted sum of the structural attributes and the valuations of the nearby homes of house \( i \). The gated attention coefficients undergo refinement during the learning phase.

\subsubsection{Final Aggregation Block}

The final aggregation stage shown in Figure \ref{fig:multihead} E involves collecting and combining the attention vectors from each head of the attention mechanism and applying the gated attention based on the normalized gates weights and biases. It is important to note that this process is unique for each attention mechanism, namely Geo and Euclidean, and it results in the formation of two separate aggregated attention vectors.

To ensure the effectiveness of the attention mechanism in both Geo and Euclidean interpolation, it is crucial to normalise the gating weights and biases using a softmax function, as shown in Equation \ref{eq:softmax_gate}. By normalizing the gating weights and biases, they fall within the range of 0 to 1, which makes them more easily interpretable.

\begin{equation}
\text{gate}_{\text{norm}, i} = \frac{\exp(\text{Gate\_weights}_i + \text{Gate\_bias}_i)}{\sum_{j=1}^{n} \exp(\text{Gate\_weights}_j + \text{Gate\_bias}_j)}
\label{eq:softmax_gate}
\end{equation}

After normalising the gating weights and biases, we perform element-wise multiplication with each attention and then aggregate them. 
The resulting vector that shows the aggregated gated geographic attention, denoted as \( v_{\text{agg\_ggeo}} \), is presented in Equation \ref{geo_agg}.

\begin{equation}
v_{\text{agg\_ggeo}} = \sum_{i=1}^{n} \text{gate}_{\text{norm, geo}, i} \odot v_{\text{ggeo}, i}
\label{geo_agg}
\end{equation}

Where \( \text{gate}_{\text{norm, geo}, i} \) represents the softmax-normalized gating weights and biases, and \( v_{\text{geo}, i} \) refers to the attention vectors from the Geo attention heads.

In a similar vein, the aggregated gated Euclidean attention vector \( v_{\text{agg\_geuc}} \) is represented by Equation \ref{euc_agg}:

\begin{equation}
v_{\text{agg\_geuc}} = \sum_{i=1}^{n} \text{gate}_{\text{norm, euc}, i} \odot v_{\text{geuc}, i}
\label{euc_agg}
\end{equation}

Here, \( \text{gate}_{\text{norm, euc}, i} \) signifies the softmax-normalized gating weights, and \( v_{\text{geuc}, i} \) portrays the gated attention vectors emergent from the Euclidean attention heads.

In conclusion, the consolidated Geo attention vector \( v_{\text{agg\_ggeo}} \) and the Euclidean attention vector \( v_{\text{agg\_geuc}} \) are computed using an element-wise multiplication between the softmax-normalized gating weights and their corresponding attention vectors as illustrated in Figure \ref{fig:multihead} F derived from the Geo and Euclidean attention heads, respectively. This approach ensures an accurate integration of the significance associated with each feature and reflects the complex spatial relationships inherent within the Geo and Euclidean contexts.

\subsection{House embeddings}
Embeddings serve as a pivotal component in contemporary machine-learning architectures, especially in scenarios that involve the manipulation of high-dimensional or categorical variables. In the realm of real estate price prediction, the utility of embeddings is accentuated for the encoding of categorical attributes such as neighbourhood classifications, types of properties, and associated amenities into a continuous vector space \cite{HouseEmbeddings}. These continuous embeddings can capture intricate relationships between disparate categories, thereby augmenting the predictive efficacy of the machine learning model \cite{MikolovWord2Vec, GloVe}. The transformation from a sparse, high-dimensional feature space to a dense, lower-dimensional vector space has found applications across a multitude of domains, ranging from natural language processing to recommendation engines and graph-based machine learning algorithms \cite{BERT, ELMO, MatrixFactorization, GraphEmbeddings}. However, effectively utilising embeddings necessitates meticulous tuning and validation to mitigate the risk of overfitting and ensure robust generalisation on unseen data \cite{OverfittingEmbeddings}.
In the present study, as delineated in Figure~\ref{fig:mylabel}, we introduce a novel methodology for generating house embeddings. Initially, two distinct Multi-Head Gated Attention mechanisms are employed: one geographically oriented (Geo Multi-Head Gated Attention) and another focused on structural attributes (Euc Multi-Head Gated Attention). The Geo Multi-Head, Gated Attention mechanism leverages the geographical coordinates of proximate properties, while the Euc Multi-Head Gated Attention mechanism utilises the structural attributes of neighbouring properties. The vectors generated from these attention mechanisms are concatenated with the original geographical (\( Gi \)) and structural (\( Ai \)) attributes of the property. This concatenated vector is propagated through a hidden neural layer to synthesise the final house embeddings. This intricate methodology enables the model to assimilate both geographical and structural nuances, thereby enhancing its predictive capabilities.

\subsection{Regression layer}
For the empirical component of our study, we employed a diverse set of regression algorithms, each optimised through rigorous cross-validation techniques. The algorithms were selected based on their suitability for the specific characteristics of our dataset as well as the computational resources at our disposal. Below is an exhaustive list of the algorithms utilised:

\begin{itemize}
    \item \textbf{Linear Regression (LR)}: Utilized with default hyperparameters as implemented in the scikit-learn library \cite{Pedregosa2011}. This algorithm serves as a baseline model for our study.
    
    \item \textbf{Random Forest (RF)}: An ensemble of decision trees, optimised using grid search and k-fold cross-validation. Hyperparameters such as the number of trees was varied, with tests conducted for 50, 100, 200, 700, and 1000 trees \cite{Breiman2001}.
    
    \item \textbf{LightGBM (LGBM)}: A gradient boosting framework that uses tree-based learning algorithms. Hyperparameters including the number of trees (50, 100, 200), the number of leaves (3, 4, 5, 100, 300), and the learning rate (0.03, 0.05, 0.07, 0.1) were fine-tuned \cite{Ke2017}.
    
    \item \textbf{Extreme Gradient Boosting (XGB)}: An optimised distributed gradient boosting library, fine-tuned through cross-validation. Parameters such as minimum child weight, gamma, subsample, column sample by the tree, learning rate, and maximum depth were adjusted \cite{chen2016xgboost}.
    
    \item \textbf{Categorical Boosting (CatBoost)}: An algorithm specifically designed for handling categorical variables. The depth parameter was optimised, with tests conducted for depths of 8 and 10 \cite{Prokhorenkova2018}.
    
    \item \textbf{K-Nearest Neighbors (KNN)}: A distance-based algorithm, optimised by adjusting the number of neighbours, with tests conducted for 10 and 15 neighbours \cite{Cover1967}.
    
    \item \textbf{Decision Tree (DT)}: A basic tree-based model, optimised by adjusting the maximum depth parameter, with tests conducted for a depth of 9 \cite{Quinlan1986}.
    
    \item \textbf{Support Vector Machines (SVM)}: A kernel-based algorithm suitable for linear and non-linear problems. Parameters 'C' and 'gamma' were fine-tuned using cross-validation \cite{Cortes1995}.
    
    \item \textbf{Regression Layer (RL)}: This layer serves as the terminal component of our attention-based neural network model, generating the final housing price prediction based on the feature map (house embeddings) obtained from preceding layers.
\end{itemize}

This empirical analysis aims to provide a comprehensive evaluation of the selected algorithms, thereby elucidating the relative merits and demerits in the context of housing price prediction.

\section{Experimentation}\label{sec7}
This section presents the experimentation methodology adopted for our house price prediction task, including the specifics of the dataset preparation, model implementation, training, and evaluation process.
\subsection{Dataset}\label{sec8}
In the experimental section, we utilised several datasets from different cities across various parts of the world to showcase the effectiveness of our model. 

\begin{table*}[htp]
\centering
\caption{Summary of Datasets}
\begin{tabular}{|l|c|c|c|}
\hline
\textbf{Dataset} & \textbf{Price Range} & \textbf{Number of Samples} & \textbf{Number of Features} \\
\hline
IT (Italian) & (60000 to 720000) Euro & 30,918 & 24 \\
\hline
BJ (Beijing) & (5500 to 170000) Yuan &  28,550 & 26 \\
\hline

KC (Kings county) & (75,000 to 7,700,000) Dollar &  21,650 & 18 \\
\hline

POA (Porto Alegre City) & (70,000 to 1,168,324) Reais &  15,368 & 7 \\
\hline
\end{tabular}
\label{tab3}
\end{table*}
\begin{enumerate}
    \item \textbf{Italian (IT) Dataset}: 
    We obtained our dataset of Italian (IT) properties from Immobiliare. It is a well-known real estate platform in Italy. To collect the data, we designed a web scraper that extracted information from eight different cities: Genoa, Milan, Turin, Rome, Bologna, Florence, Naples, and Palermo. We filtered the data to include only five types of properties, such as apartments and penthouses while excluding outliers like farms, buildings, and properties under construction. This ensured that the dataset was representative and coherent.
    We then conducted a thorough cleaning process to eliminate outliers. This process helped us eliminate data entry errors and rare property types, resulting in a consistent dataset suitable for analysis. To enrich the dataset, we added geographical data points. We included precise longitude and latitude coordinates for each property listing and leveraged OpenStreetMap to enhance each listing with Points of Interest (POI) data. This provided more profound insights into the property's surroundings, which could be significant in assessing its value.
    The final IT dataset comprises 30,918 property listings spread across eight significant cities in Italy. Each listing includes 19 distinct features that capture structural attributes, such as surface area, year of construction, and geographical details. 

    \item \textbf{Beijing (BJ) Dataset}: This dataset consists of 28,550 real estate transactions in Beijing and is sourced from the H4M study \cite{zhao2022h4m}. It includes 25 features, which range from structural attributes like surface area and year of construction to geographical elements such as district location and Point Of Interest (POI) information. The features are detailed in Table \ref{tab3}. 

    \item \textbf{Kings County (KC) Dataset}: Sourced from the GitHub\footnote{\url{https://github.com/darniton/ASI}} In the repository associated with the "Attention-Based Interpolation" paper, there is a dataset representing the Kings County, USA housing market. This dataset comprises 21,650 house samples, characterized by 19 distinct features. These features, which encompass both structural and geographical attributes, are detailed in a separate table, Table \ref{tab3}.

    It's important to note that the prices in this dataset are provided in a log-scaled format.

    \item \textbf{Porto Alegre City (POA) Dataset}: Derived from the repository provided by Vianna and Barbosa, this dataset focuses on Brazil's Porto Alegre City housing market. It includes 15,368 house samples, each described by 6 features, similar to the KC dataset. The features are outlined in Table \ref{tab3}.

    It's essential to recognize that the prices in this dataset are provided in a log-scaled format.

\end{enumerate}

\subsection{Model configuration}\label{sec9}
Our model was developed in a Python 3.7 environment, using TensorFlow 2.5 as the backend for the Keras framework. The model was executed on a system with an Intel Core i5-13700K CPU and an NVIDIA GeForce RTX 3070 GPU. We used cross-validation and grid search techniques for hyperparameter tuning to achieve optimal results with regression algorithms such as XGBoost and RandomForest. For our custom model, we fine-tuned the hyperparameters using a validation subset of the data to obtain the best possible embedding representation and predictive performance. The hyperparameters and their values are summarised in Table~\ref{tab:hyperparameters}, and we describe each hyperparameter and its significance below.
\begin{itemize}
      \item \textbf{n-nearest}: Specifies the number of nearest neighbours to consider. The best values were 40 for IT, 60 for KC, 60 for POA, and 30 for BJ.
    
      \item \textbf{sigma ($\sigma$)}: Controls the width of the Gaussian kernel. Optimal values were 2 for IT, 2 for KC, 2 for POA, and 10 for BJ.
    
      \item \textbf{nodes}: Represents the number of nodes in the hidden layers. The best values were 60 for IT, 60 for KC, 60 for POA, and 60 for BJ.
      
      \item \textbf{Num\_heads}: Specifies the number of attention heads in the model. Optimal values were 8 for IT, 8 for KC, 4 for POA, and 4 for BJ.
    
      \item \textbf{num\_geo}: Indicates the number of geographical features to consider. The best values were 30 for IT, 30 for KC, 10 for POA, and 15 for BJ.
    
      \item \textbf{num\_euc}: Represents the number of Euclidean dimensions for distance calculations. The best values were 25 for IT, 30 for KC, 15 for POA, and 15 for BJ.
      
      \item \textbf{LR (Learning Rate)}: Controls the step size during optimization. Optimal values were 0.001 for IT, 0.008 for KC, 0.001 for POA, and 0.001 for BJ.
    
      \item \textbf{batch size}: Specifies the number of samples per batch during training. Optimal values were 32 for IT, 250 for KC, 32 for POA, and 250 for BJ.
    
      \item \textbf{act func (Activation Function)}: Either Rectified Linear Unit (ReLU) or Exponential Linear Unit (ELU) was used. ELU was optimal for all datasets.
    
      \item \textbf{hidden act function (Hidden Layer Activation Function)}: The activation function for the hidden layers was either ReLU, ELU, regression, or linear. The linear function was optimal for all datasets.
      
      \item \textbf{similarity function}: We used the Gaussian Kernel and Identity function to compute similarities between data points. The Identity function was optimal for IT and POA, while the Gaussian Kernel was optimal for KC and BJ.
\end{itemize}

\begin{table}[h]
\centering
\caption{hyperparameters that were used to train our model}
\begin{tabular}{|c|c|c|c|c|c|}
\hline
\multirow{2}{*}{HP Values} & \multirow{2}{*}{General Values} & \multicolumn{4}{c|}{Best Values} \\

& & IT & KC & POA & BJ  \\
\hline
N-nearest & 5, 10, 15, 60 & 40 & 60 & 60 & 30  \\
\hline
Nearest-geo & 20, 25, 30, 35, 40, 45, 50, 55, 60 & 30 & 30 & 10 & 15 \\
\hline
Nearest-Euclid & 20, 25, 30, 35, 40, 45, 50, 55, 60 & 25 & 30 & 15 & 15 \\
\hline
Num\_heads & 1,2,4,8,12,15 & 8 & 8 & 4 & 4 \\
\hline
Sigma($\sigma$) & 2, 5, 10, 15, 20 & 2 & 2 & 2 & 10 \\
\hline
Nodes & 5, 10, 15, 60 & 60 & 60 & 60 & 60 \\
\hline
LR & [0.001-0.01] & 0.001 & 0.008 & 0.001 & 0.001  \\
\hline
Batch size & 250, 300, 400, 500 & 32 & 250 & 32 & 250 \\
\hline
Act func & Relu and ELU & ELU & ELU & ELU & ELU  \\
\hline
Hidden act func & Relu, ELU, regression and linear & linear & linear & linear & linear \\
\hline
Similarity function & Identity and Gaussian Kernel & Identity & Gaussian Kernel & Identity & Gaussian Kernel \\
\hline
\end{tabular}
\label{tab:hyperparameters}
\end{table}
\subsubsection{Evaluation Metrics}\label{sec9}
After training, the model was evaluated using standard regression metrics such as Root Mean Squared Error (RMSE) and Mean Absolute Error (MALE). These metrics serve specific purposes in assessing the model's performance:

\begin{itemize}
    \item \textbf{RMSE (Root Mean Squared Error)}: Provides a measure of the model's prediction error, penalising more significant errors more severely than smaller ones. It is advantageous when significant errors are undesirable in the prediction task.
    
    \item \textbf{MALE (Mean Absolute Logarithmic Error)}: This metric expresses the average magnitude of the relative errors between predicted and actual values while disregarding their direction. It is beneficial when dealing with exponential growth, or underestimation is more critical than overestimation.
\end{itemize}

These metrics collectively offer a comprehensive evaluation of the model's performance in predicting house prices, allowing for the assessment of the model's accuracy and its goodness of fit to the actual data.

\subsection{Results and Interpretation}\label{sec10}
In our evaluation, we consider both the average and best performance metrics to offer a comprehensive view of each model's capabilities. The average performance metrics are derived from 10-fold cross-validation, indicating how the model will likely perform on unseen data. It gives us a more generalisable performance measure by mitigating the risk of the model overfitting to a particular subset of the data. On the other hand, the best performance metrics are extracted using grid search techniques. These values demonstrate the optimal performance that the model can potentially achieve under ideal hyperparameter settings. Including both types of metrics allows for a balanced understanding of the model's robustness and potential for excellence. It helps in identifying not just the most consistently high-performing models but also those with the capacity for superior performance under the right conditions.
\subsubsection{Base models benchmark}\label{sec11}
Table~\ref{base_benchmark} provides an exhaustive evaluation of multiple machine-learning models 
In an exhaustive evaluation of machine learning models on real estate datasets from Italy (IT), King's County (KC), Porto Alegre City in Brazil (POA), and Beijing (BJ), the best performance was consistently demonstrated by XGBoost (XGB). Specifically, XGB recorded the best MALE values of 0.1350 in IT, 0.1160 in KC, 0.1613 in POA, and 0.0723 in BJ. Notably, the average performance for XGB was stable and closely aligned with these best values, indicating high reliability across diverse geographic datasets. CatBoost and LightGBM also performed strongly, closely trailing XGB in each dataset. For instance, CatBoost had the best MALE values of 0.1362 in IT, 0.1131 in KC, 0.1793 in POA, and 0.0782 in BJ. LightGBM posted the best MALE deals of 0.1381 in IT, 0.1164 in KC, 0.172 in POA, and 0.0790 in BJ. The average performances of CatBoost and LightGBM were also impressively stable and nearly matched their respective best values.
Conversely, Support Vector Machines (SVM) significantly underperformed, with its best MALE values being 0.4072 in IT, 0.1331 in KC, 0.2232 in POA, and a dismal 0.2234 in BJ. K-Nearest Neighbors (KNN), a traditional algorithm, also lagged, particularly in the BJ dataset, where it posted a best MALE of 0.1116. In summary, XGB takes the lead across all datasets regarding best and average performance metrics, closely followed by CatBoost and LightGBM, which also show highly stable average performances. Conversely, SVM and traditional models like KNN are less effective, particularly in complex, geographically diverse datasets.

\begin{table}

    \centering
    
    \caption{Benchmark the datasets on state-of-the-art machine learning models. The average value is referred to k-fold cross-validation with k=10}
        \centering
    \begin{tabular}{l@{\hspace{4pt}}c@{\hspace{2pt}}c@{\hspace{2pt}}c@{\hspace{2pt}}c@{\hspace{2pt}}c@{\hspace{2pt}}c@{\hspace{2pt}}c@{\hspace{2pt}}c@{\hspace{2pt}}c@{\hspace{2pt}}c@{\hspace{2pt}}c@{\hspace{2pt}}c@{\hspace{2pt}}c@{\hspace{2pt}}c@{\hspace{2pt}}c@{\hspace{2pt}}c}
        \hline
        Model & \multicolumn{4}{c}{IT} & \multicolumn{4}{c}{KC} & \multicolumn{4}{c}{POA} & \multicolumn{4}{c}{BJ} \\
        & \multicolumn{2}{c}{MALE $\downarrow$} & \multicolumn{2}{c}{RMSE $\downarrow$} & \multicolumn{2}{c}{MALE $\downarrow$} & \multicolumn{2}{c}{RMSE $\downarrow$} & \multicolumn{2}{c}{MALE $\downarrow$} & \multicolumn{2}{c}{RMSE $\downarrow$} & \multicolumn{2}{c}{MALE $\downarrow$} & \multicolumn{2}{c}{RMSE $\downarrow$} \\
        & Best & Avg & Best & Avg & Best & Avg & Best & Avg & Best & Avg & Best & Avg & Best & Avg & Best & Avg \\
        \hline \\
        LR & 0.385 &  0.388 & 76224 & 76241 & 0.1924 & 0.1925 & 205460 & 209330 & 0.2610 & 0.2611 & 152861 & 153775 & 0.2394 & 0.2396 & 20551 & 20558 \\\\
        
        KNN & 0.247 & 0.248 & 84637 & 85163 & 0.1501 & 0.1513 & 174046 & 175628 & 0.2065 & 0.2078 & 122521 & 123113 & 0.1116 & 0.1121 & 12093 & 12149 \\\\
        
        DT & 0.197 & 0.205 & 69085 & 70423 & 0.1583 & 0.1608 & 158937 & 178296 & 0.2163 & 0.2195 & 127382 & 128915 & 0.0936 & 0.0954 & 10155 & 10409 \\\\
        
        RF & 0.1502 & 0.1508 & 51774 & 52147 & 0.1245 & 0.1251 & 133933 & 136993 & 0.1716 & 0.1731 & 105183 & 105975 & 0.0784 & 0.0794 & 8369 & 8475 \\\\
        
        SVM &  0.4072 & 0.4074 & 128634 & 128781 & 0.1331 & 0.1336 & 149265 & 152675 & 0.2232 & 0.2246 & 126911 & 128191 & 0.2234 & 0.2237 & 20652 & 20668 \\\\
        
        LGBM & 0.1381 & 0.1384 & 46183 & 46492 & 0.1164 & 0.1175 & 122116 & 126076 & 0.172 & 0.177 & 104928 & 106705 & 0.0790 & 0.0796 & 8070 & 8152 \\\\
        
        CatBoost & 0.1362 & 0.1368 & \textbf{45942} & \textbf{46233} & 0.1131 & 0.1141 & 120351 & \textbf{123077} & 0.1793 & 0.1775 & 105984 &  106593 & 0.0782 & 0.0785 & 7995 & 8066 \\\\
        
        XGB & \textbf{0.1350} & \textbf{0.1358} & 46008 & 46396 & \textbf{0.1160} & \textbf{0.1167} & \textbf{119479} & 124459 & \textbf{0.1613} & \textbf{0.1634} & \textbf{100212} & \textbf{101614} & \textbf{0.0723} & \textbf{0.0744} & \textbf{7713} & \textbf{7836} \\
        \hline
    \end{tabular}

    \label{base_benchmark}

\end{table}

\subsubsection{Experimental results for our model}\label{sec11}
We conducted a series of tests to evaluate our model and compare it with ANN and ASI models. The dataset was split into three subsets, with 70\% assigned for training, 20\% for testing, and 10\% for validation. Table~\ref{asi_preformance} shows the performance metrics of our model, ASI, and ANN, compared across four different datasets from various locations: Italy (IT), King's County (KC), Porto Alegre (POA), and Beijing (BJ).

Our model performed better than the ASI model in the IT dataset. It has a superior MALE that's approximately 1.52\% lower and an RMSE that's approximately 0.36\% lower, with figures of 0.1312 and 45,797, respectively. These results highlight our model's ability to interpret the IT dataset more accurately.

In the KC dataset, our model's robustness shines with a notable 13.3\% improvement in RMSE compared to the ASI model. This translates to a lower RMSE of approximately 107,993 for our model compared to ASI's 124,557. This underscores our model's consistent efficiency across various geographical contexts.

Our model also outperforms the ASI model in the POA and BJ datasets, demonstrating its versatility and accuracy in handling diverse data types. In the POA dataset, our model achieves a MALE of approximately 1.44\% lower and an RMSE of approximately 13.45\% lower compared to ASI. In the BJ dataset, our model achieves a MALE of approximately 2.67\% lower and an RMSE of 2.31\% lower compared to ASI. These metrics align with those observed in the IT and KC datasets, reaffirming the broad applicability of our model.

Our model incorporates Multi-Head Gated Attention mechanisms, which enable it to assimilate diverse spatial cues between houses and their surroundings. This fosters more precise and robust predictions as our model comprehensively grasps spatial dynamics.

In conclusion, our model has proven superior to the ASI model across all evaluated metrics and datasets. The advanced Multi-Head Gated Attention architecture plays a pivotal role in aggregating contextual cues, ultimately enhancing overall predictive accuracy.

\begin{table*}
    \centering
    \caption{Performance Evaluation of our model against the ASI model and ANN}
    \begin{tabular}{l@{\hspace{2pt}}c@{\hspace{2pt}}c@{\hspace{2pt}}c@{\hspace{2pt}}c@{\hspace{2pt}}c@{\hspace{2pt}}c@{\hspace{2pt}}c@{\hspace{2pt}}c}
        \hline
        Model & \multicolumn{2}{c}{IT} & \multicolumn{2}{c}{KC} & \multicolumn{2}{c}{POA} & \multicolumn{2}{c}{BJ} \\\
        & MALE $\downarrow$ & RMSE $\downarrow$ & MALE $\downarrow$ & RMSE $\downarrow$ & MALE $\downarrow$ & RMSE $\downarrow$ & MALE $\downarrow$ & RMSE $\downarrow$ \\
        \hline\\
        ANN & 0.197 & 67835 & 0.2231 & 127900 & 0.2212 & 125961 & 0.239 & 19565 \\\\
        ASI & 0.133 & 46473 & 0.112 & 124557 & 0.139 & 93818 & 0.075 & 7934 \\\\
        Ours & \textbf{0.1312} & \textbf{45797} & \textbf{0.110} & \textbf{107993} & \textbf{0.136} & \textbf{92020} & \textbf{0.073} & \textbf{7797} \\
        \hline
    \end{tabular}
    \label{asi_preformance}
\end{table*}

\subsubsection{Experimental results for house embeddings}\label{sec11}
In our experiment, we wanted to see how custom house embeddings generated by our Multi-Head Gated Attention model would affect the performance of various baseline machine learning models. These embeddings were created based on structural and geographical information and enhanced the feature space for algorithms like Linear Regression, KNN, Decision Tree, Random Forest, SVM, LightGBM, CatBoost, and XGBoost. We evaluated the models' performance using four different geographical datasets: Italy (IT), King's County (KC), Porto Alegre (POA), and Beijing (BJ), and assessed the Best and Average MALE and RMSE scores.

Our results showed that our custom embeddings significantly positively impacted the predictive performance of the baseline models. For example, when the CatBoost model was augmented with our custom embeddings, it achieved the lowest RMSE score in the IT dataset at 45,708, outperforming even our original Multi-Head Gated Attention model. However, we found that the improvement magnitude was inconsistent across all datasets. The IT dataset, which combines data from various cities with significant geographical and Euclidean distances between them, showed only a modest enhancement of around 1.3\% in RMSE when deploying CatBoost with custom embeddings compared to the baseline. 

We discovered that the unique spatial complexities inherent in each dataset could impact the effectiveness of the custom embeddings. For instance, in the KC dataset, CatBoost with custom embeddings demonstrated significant gains over its baseline, whereas, in IT, the improvements were more restrained. We also found that even simpler models like Linear Regression could benefit substantially from the enriched feature space the embeddings provide. In the IT dataset, the best MALE improved by approximately 65.8\%, the average MALE improved by approximately 66.0\%, the best RMSE improved by approximately 39.8\%, and the average RMSE improved by approximately 39.8\%.

In the CatBoost model for the IT dataset, the best MALE improved by approximately 2.4\%, and the average MALE improved by approximately 2.9\%. The best RMSE improved by approximately 0.5\%, and the average RMSE improved by approximately 1.0\%. This indicates a positive trend in reducing MALE and RMSE values, which is crucial for achieving better model performance in predictive tasks like house price prediction.

In the KC dataset, the implementation of custom embeddings reflected varying degrees of improvement across different machine-learning models. The CatBoost model illustrated an enhancement in the best MALE value by approximately 5\%, although the average MALE value experienced a minor deterioration by approximately 0.44\%. On the brighter side, a more noticeable improvement was observed in the RMSE values, where the best RMSE value improved by approximately 8.20\%, and the average RMSE value improved by approximately 8.04\%.

The POA dataset manifested a significant leap in performance metrics upon integrating custom embeddings. Specifically, the CatBoost model, when augmented with custom embeddings, demonstrated a robust improvement in both MALE and RMSE values. The best MALE value improved by an impressive margin of approximately 23.77\%, while the average MALE value improved by approximately 22.66\%. Concurrently, the RMSE metrics also exhibited substantial enhancements, with the best RMSE value improving by approximately 13.40\%, and the average RMSE value improving by approximately 13.45\%.

In the BJ dataset, we observed that models trained on embeddings generally perform better on average values, reflecting a more consistent performance across varying data points. However, the best values achieved in MALE and RMSE metrics were slightly better when models were trained on original data. This suggests that while embeddings generally enhance model performance, there might be specific instances or datasets where traditional feature sets could yield better or comparable results.

\begin{table}
    \centering
    \caption{Benchmark the datasets on state-of-the-art machine learning models on the generated embeddings from our model. The average value is referred to k-fold cross-validation with k=10}
    \centering
    \begin{tabular}{l@{\hspace{4pt}}c@{\hspace{2pt}}c@{\hspace{2pt}}c@{\hspace{2pt}}c@{\hspace{2pt}}c@{\hspace{2pt}}c@{\hspace{2pt}}c@{\hspace{2pt}}c@{\hspace{2pt}}c@{\hspace{2pt}}c@{\hspace{2pt}}c@{\hspace{2pt}}c@{\hspace{2pt}}c@{\hspace{2pt}}c@{\hspace{2pt}}c@{\hspace{2pt}}c}
        \hline
        Model & \multicolumn{4}{c}{IT} & \multicolumn{4}{c}{KC} & \multicolumn{4}{c}{POA} & \multicolumn{4}{c}{BJ} \\
        & \multicolumn{2}{c}{MALE $\downarrow$} & \multicolumn{2}{c}{RMSE $\downarrow$} & \multicolumn{2}{c}{MALE $\downarrow$} & \multicolumn{2}{c}{RMSE $\downarrow$} & \multicolumn{2}{c}{MALE $\downarrow$} & \multicolumn{2}{c}{RMSE $\downarrow$} & \multicolumn{2}{c}{MALE $\downarrow$} & \multicolumn{2}{c}{RMSE $\downarrow$} \\
        & Best & Avg & Best & Avg & Best & Avg & Best & Avg & Best & Avg & Best & Avg & Best & Avg & Best & Avg \\
        \hline\\
        LR & \textbf{0.1317} & \textbf{0.1318} & 45837 & 45868 & \textbf{0.1103} & \textbf{0.1104} & \textbf{106954} & \textbf{107133} & 0.1369 & 0.1372 & 91725 & 91848 & \textbf{0.0732}  & \textbf{0.0733} & \textbf{7779} & \textbf{7786} \\\\
        KNN & 0.1352 & 0.1354 & 46648 & 46761 & 0.1208 & 0.1211 & 123235 & 125010 & 0.1398 & 0.1402 & 92032& 92369  &0.0767 &  0.0770 & 7980  & 8015 \\\\
        DT & 0.1347 & 0.135 & 46007 & 46160 & 0.1353 & 0.1412 & 142731 & 154575 & 0.1501 & 0.1512 & 97704 & 98387 & 0.0752 &  0.0756 & 7853 & 7879 \\\\
        RF & 0.1323 & 0.1325 & 45921 & 45995 & 0.1146 & 0.1151 & 112588 & 115852 & 0.1388 & 0.1396 & 92623 &93243 & 0.0741 &  0.0742 & 7843 & 7864 \\\\
        SVM & 0.1743 & 0.1745 & 58977 & 59188 & 0.1103 & 0.1103 & 107389 & 108700 & \textbf{0.1357} & \textbf{0.1359} & \textbf{91719} & \textbf{91796} & 0.0778  & 0.0779 & 8332 & 8344\\\\
        LGBM & 0.1324 & 0.1327 & 45885 & 45930 & 0.1138 & 0.1141 & 111551 & 112994 & 0.1384 & 0.1387 & 92383 & 92617 &0.0742  & 0.0744& 7835 &7854 \\\\
        CatBoost &0.1320 & 0.1321 & \textbf{45708} & \textbf{45752} & 0.1130 & 0.1136 & 110481 & 113174 & 0.1367 & 0.1373 & 91784 & 91976 & 0.0735 & 0.0737 & 7806 & 7814 \\\\
        XGB & 0.1324 & 0.1325 & 45961 & 46021 & 0.1147 & 0.1152 & 108644 & 112332 & 0.1392 & 0.1397 & 92677  & 93003 &  0.0739 & 0.0740
        & 7822  & 7834 \\
        \hline
    \end{tabular}
 \label{base_benchmark_emb}
\end{table}
\subsection{Discussion}
\label{subsec:benchmarking}

\label{subsec: superiority}Building upon our previous results in Table \ref{base_benchmark},\ref{asi_preformance},\ref{base_benchmark_emb} our model, based on Multi-Head Gated Attention, consistently outperforms the baseline models across multiple datasets. This superiority is particularly noteworthy as the model excels in spatial interpolation tasks and enhances the performance of other state-of-the-art machine learning models when its embeddings are used. One of the key advantages of our model over the attention-based interpolation model is the ability to capture multiple contexts from each head and control the flow of the information so that it will consider the most similar neighbours through the use of Multi-Head Gated Attention.

\subsubsection{Spatial and Structural Analysis}

The present study introduces a Multi-Head Gated Attention model that exhibits superior performance compared to baseline models when applied to various datasets, including IT and POA. This model utilizes distinct weights and biases within each attention head to capture various contextual relationships within the data, showcasing its exceptional capabilities in spatial interpolation tasks. This approach provides a more comprehensive understanding of the underlying spatial dynamics. Our model's multi-head gated attention mechanism exceeds traditional singular attention approaches by integrating various spatial and structural features from the data. This integration is essential as it moderates the influence of outliers, which is expected in a vast and diverse metropolis like Beijing, where extreme data points can skew the analysis. By employing this sophisticated mechanism, the model ensures the delivery of accurate and nuanced house price predictions that genuinely reflect the complex intricacies of Beijing's housing market, setting a new benchmark for robustness and reliability.
The box plots in Figure \ref{fig:enter-label_graphbox}(a,b) effectively illustrate each dataset's spatial and structural features. Specifically, Figure \ref{fig:enter-label_graphbox}(b) reveals that Kings County (KC) has a compact urban form, indicated by a median geodesic distance of just under 0.65 km, which is also supported by a low median normalized Euclidean distance shown in Figure \ref{fig:enter-label_graphbox}(a), highlighting high structural homogeneity among houses.
In contrast, Beijing (BJ) portrays a more dispersed housing structure with a median geodesic distance of approximately 0.45 km, as indicated in Figure \ref{fig:enter-label_graphbox}(b), and a median normalized Euclidean distance of approximately 0.150, as shown in Figure \ref{fig:enter-label_graphbox}(a). These distances indicate a significant variation in structural features, suggesting a housing landscape that includes densely packed urban areas and more spread-out suburban or peri-urban zones. The Italian (IT) region demonstrates a median geodesic distance of around 0.50 km, reflecting less uniformity and greater architectural diversity, as further evidenced by a median normalized Euclidean distance of around 0.110.
Moving to Porto Alegre (POA), the dataset displays a distinctive spatial composition, with a median geodesic distance that suggests moderately dense housing and a median normalized Euclidean distance of approximately 0.100. This places POA in a unique position between the densely packed environment of KC and the varied spatial arrangements of BJ and IT. The moderate variation in POA's housing structures signifies an urban design that merges densely built areas with open suburban spaces, reflecting its rich historical development and cultural diversity.
Employing the multi-head gated attention mechanism for the POA dataset allows for an in-depth exploration of the city's complex architectural styles and spatial dynamics. When juxtaposed with the consistent architecture of KC and the diverse spatial distributions of BJ and IT, our model's multifaceted approach yields a deep understanding of the nuances within POA's urban clusters and the distinctive nature of its rural homes. As a result, our model stands out as a sophisticated and precise analytical tool, uniquely equipped to navigate and predict the intricate dynamics of the housing market with extraordinary accuracy and insight.

The improvements highlighted in Table 
\ref{asi_preformance} emphasises the progress made by our model compared to the ASI model. Our model achieved improvements of 1.35\% and 1.46\% in MALE and RMSE, respectively, for the IT dataset, 1.79\% and 13.34\% for the KC dataset, 2.16\% and 1.92\% for the POA dataset, and 2.67\% and 1.73\% for the BJ dataset. These results demonstrate the superiority of our model across different datasets and spatial configurations. The multi-head gated attention mechanism played a significant role in achieving these improvements. It captures diverse contextual relationships within the data by leveraging weights and biases in each head, especially when dealing with regions with a more varied architectural landscape and pronounced geographical diversity. The improvements in the KC dataset are significant, as it has a high degree of architectural uniformity. However, the model could still capture minute differences and nuances, leading to a 13.34\%  improvement in RMSE. For the BJ dataset, which has a more dispersed housing layout and a vast spatial range, the model achieved a 2.67\% improvement in MALE and a 1.73\% improvement in RMSE, highlighting the model's ability to accurately capture the essence of each area despite the considerable differences in spatial dynamics and architectural styles. The advances in the IT dataset were also noteworthy, with the model achieving a 1.35\% improvement in MALE and a 1.46\% improvement in RMSE despite the unique spatial layout of the region compared to KC. These results demonstrate the robustness and reliability of our model in providing accurate predictions against the ASI model across diverse datasets and spatial configurations.

\subsubsection{Emebeddings performance}
\label{subsec:effectiveness_HE}

In Table \ref{base_benchmark_emb}, we present a comparative analysis of our model embeddings against the benchmarks outlined in Table \ref{base_benchmark}. Additionally, the results from the regression layer of our model are presented in Table \ref{asi_preformance}. The results underline the substantial advancements made by our model and the generated embeddings. Rigorous evaluations across various validation sets demonstrate the superior performance of our model in handling complex spatial datasets. Furthermore, the efficiency of the generated embeddings emphasises our model's role in reducing data complexity so that simple models like linear regression can outperform ensembling models.

In the IT dataset, our model achieved a Mean Absolute Logarithmic Error (MALE) of 0.1312 and a Root Mean Square Error (RMSE) of 45,797. These results represent a 2.89\% improvement in MALE and a 0.46\% improvement in RMSE over the best baseline model, XGBoost, which recorded a MALE of 0.1350 and an RMSE of 46,008.

Furthermore, the embeddings in our model outperformed the regression layer of our model and the base benchmarking in terms of RMSE, with the Catboosting model achieving the best result of 45,708. This indicates a slight improvement over our model's performance.

\begin{figure*}
    \centering
    \begin{minipage}{0.48\textwidth}
        \centering
        \subfloat[]{%
            \includegraphics[width=\textwidth]{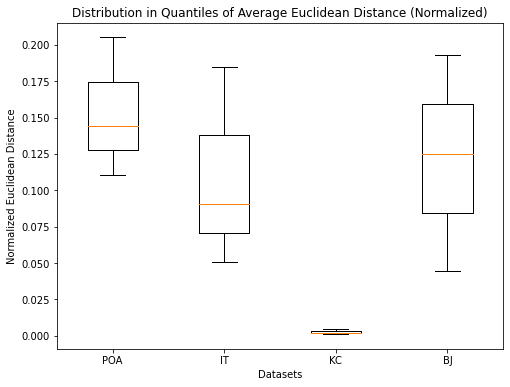}%
            \label{fig:subfig_a}%
        }
    \end{minipage}
    \hfill
    \begin{minipage}{0.48\textwidth}
        \centering
        \subfloat[]{%
            \includegraphics[width=\textwidth]{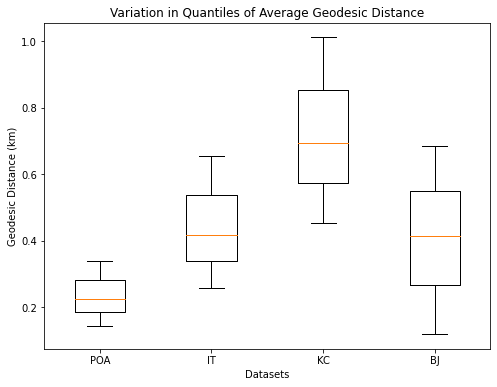}%
            \label{fig:subfig_b}%
        }
    \end{minipage}
\caption{Analysis of Geodesic and Euclidean Distances Among the 60 Nearest Houses Across Datasets
\protect\subref{fig:subfig_a} Highlights the variation in quantiles of the average geodesic distance (in km) for the 60 nearest houses across the four datasets, reflecting the spatial proximity of residences.
\protect\subref{fig:subfig_b} Represents the distribution in quantiles of the average normalised Euclidean distance for the 60 nearest houses, taking into account the structural features of the houses. Min-max normalisation was employed to standardise the distance values due to the diverse attributes of the houses in each dataset.
}
    \label{fig:enter-label_graphbox}
\end{figure*}

These results can be attributed to the challenging nature of predicting housing prices accurately in this dataset, where various factors come into play. Our model's success suggests that its embeddings effectively capture the price variations associated with the diverse housing landscape, as evident from the wide distribution of Euclidean distances in Figure \ref{fig:enter-label_graphbox}(a). This distribution reflects the influence of different cities in one dataset, especially Italian cities, which exhibit various housing structures from the south to the north of italy.

Transitioning to the KC dataset, our model displayed a MALE of 0.110 and an RMSE of 107993. This corresponds to a percentage improvement of 2.81\% and 10.27\% in MALE and RMSE, respectively, compared to the best baseline model, CatBoost. CatBoost had a MALE of 0.1131 and an RMSE of 120351. However, the embeddings seem to mark the best results over our model, and the base benchmarking with 0.1103 MALE value and 106954 RMSE scored in the linear regression model shows an improvement in comparison to our model in both metrics, further emphasising the power of our generated embeddings.

Furthermore, the significant improvement observed in the Kings County (KC) dataset demonstrates our model's enhanced capability in dense housing and architectural uniformity regions. Our model boosts the prediction accuracy for the most relevant houses and creates diverse contextual frameworks that underscore the interrelationships between houses, even in areas of uniformity. Additionally, creating embeddings encapsulating these relationships further improves the model’s performance.

Our model exhibits exceptional performance on the Porto Alegre (POA) dataset, achieving the lowest Mean Absolute Logarithmic Error (MALE) at 0.136 and Root Mean Square Error (RMSE) at 92,020. This performance surpasses the XGBoost baseline's MALE of 0.1613 and RMSE of 100,212, indicating a 15.67\% improvement in MALE and an 8.17\% improvement in RMSE. The model's superior embeddings are instrumental in this achievement, effectively streamlining intricate urban data for linear regression without losing essential details, as indicated in Tables \ref{base_benchmark_emb} and \ref{asi_preformance}. Figures \ref{fig:enter-label_graphbox}(a,b) potentially reveal the spatial complexity of POA, with its moderately dense urban fabric intertwined with suburban and rural patches. Despite this diversity posing challenges for predictive models, our embeddings adeptly encode these complexities, effectively representing the multifaceted housing styles and values within POA. Our model's predictive precision stems from its algorithmic sophistication and nuanced understanding of the region’s unique urban tapestry.

Lastly, base benchmarking for the Beijing (BJ) dataset performs better than our model's regression layer. However, our embeddings demonstrate better results, suggesting they are more generalized than the base benchmarking outcomes. The embeddings score the best MALE of 0.072 and the best RMSE of 7,713, compared to 0.073 and 0.0732 MALE, and 7,797 and 7,779 RMSE with our model and linear regression model using our embeddings, respectively. Our embeddings' average values from cross-validation are 0.0733 MALE and 7,786 RMSE, while the base benchmarking average values are 0.074 MALE and 7,836 RMSE, showing a close similarity to the embeddings.

Examining the housing market in Beijing presents several challenges, including managing diverse and often extreme data points typical of a large metropolis. The median distance to the nearest 60 homes in Beijing, as depicted in Figure \ref{fig:enter-label_graphbox}(b), is approximately 0.45 km, highlighting an extensive and varied housing layout. The city's diverse architectural styles add another layer of complexity to the dataset. Our model, equipped with a Multi-Head Gated Attention mechanism, is adept at handling these challenges. This mechanism effectively regulates the influence of outliers, ensuring a nuanced and accurate representation of Beijing's housing landscape. The embeddings generated by our model are particularly noteworthy for their ability to generalize across Beijing's diverse housing market. While the base benchmarking results provide valuable insights, our model's embeddings capture a broader range of intricacies, ensuring they are statistically sound and meaningfully representative of the real-world scenario.

This quantitative comparison highlights the considerable enhancements of our model. The marked performance uplift in the BJ dataset accentuates our model's potential in real estate price prediction tasks. Additionally, the comparative analysis with the original attention-based interpolation model by Viana et al. \cite{viana2021attention} on the KC and POA datasets further amplifies the strengths of our model. Our model's ability to efficaciously reduce data dimensionality while retaining crucial information has led to significant improvements in MALE and RMSE across all datasets. This proficiency in compressing high-dimensional data into more digestible forms has enabled algorithms like linear regression to compete and outperform complex ensemble models like LightGBM, CatBoost, and XGBoost.

\section{Conclusions}
This study marked a significant advancement in the realm of house price prediction, particularly in the application of spatial interpolation techniques. One of the cornerstone contributions of our research was the introduction of a novel dataset focused on the Italian housing market. This dataset enriched the existing pool of resources and offered a unique landscape for testing new methodologies.
Our comprehensive review revealed a noticeable gap in applying attention mechanisms within house price prediction. Most notably, the use of Multi-Head Gated attention in spatial interpolation was virtually unprecedented, especially for datasets that were not time series. Our Multi-Head Gated Attention model successfully bridged this gap, demonstrating a marked improvement in prediction accuracy over traditional and original attention-based interpolation models. This underscored the untapped potential of attention mechanisms in capturing intricate spatial dependencies.
However, it was crucial to acknowledge the computational cost associated with our approach. The complexity of the model posed challenges for real-time applications or scenarios with limited computational resources.
As we looked to the future, our research aimed to expand the horizons of house price prediction by incorporating additional data sources. These included satellite imagery and both interior and exterior photographs of properties. Such multi-modal data integration would offer a more holistic view of the factors influencing house prices, thereby enhancing the predictive capabilities of our model.


\section{Acknowledgments}
The authors thank Mr. Arturo Argentieri from CNR-ISASI Italy for his technical contribution to the multi-GPU computing facilities.

\section{Declarations}
Author Contributions 
Conceptualization: Zakaria.A Sellam; Methodology: Zakaria.A Sellam; 
Literature search: Zakaria.A Sellam; Data
analysis: Zakaria.A Sellam ,Pier Luigi Mazzeo; Writing - original draft preparation: 
Zakaria.A Sellam ,Cosimo Distante ,Pier Luigi Mazzeo; Writing - review and editing: Cosimo Distante, Abdelmalik Taleb-Ahmed ;

\subsection{Funding acquisition}
Cosimo Distante; 

\subsection{Supervision}
Cosimo Distante, Abdelmalik Taleb-Ahmed.

\subsection{Funding}
This research was funded in part by Future Artificial Intelligence Research—FAIR CUP B53C220036 30006 grant number PE0000013, and in part by the Apulia Region with “Programma Regionale RIPARTI - assegni di RIcerca per riPARTire con le Imprese” POC PUGLIA FESR/FSE 2014/2020 grant 2caeb4ba and e6446c33.

\subsection{Data Availability} 
All data generated or analysed during this study are
included in this published article.
Declarations

\subsection{Competing interests} 
The authors have no relevant financial or non financial interests to disclose.

\subsection{Open Access}
This article is available under an open access policy. It permits use, sharing, adaptation, distribution, and reproduction in any medium or format, as long as appropriate credit is given to the original author(s) and the source. For additional resources, materials, and code related to this article, please visit our GitHub repository at \url{https://github.com/ldb0071/Boosting-House-Price-Estimations-with-Multi-Head-Gated-Attention/tree/main/ASI-main}. All users are required to adhere to these open access terms, ensuring proper acknowledgment of the original work.

\bibliography{sn-bibliography}

\end{document}